# An Energy-Efficient Ensemble Approach for Mitigating Data Incompleteness in IoT Applications

Yousef AlShehri
*School of Computing*
*University of Georgia*
Athens, United States
yousef.alshehri@uga.edu

Lakshmish Ramaswamy
*School of Computing*
*University of Georgia*
Athens, United States
laksmr@uga.edu

*Abstract*—Machine Learning (ML) is becoming increasingly important for IoT-based applications. However, the dynamic and ad-hoc nature of many IoT ecosystems poses unique challenges to the efficacy of ML algorithms. One such challenge is data incompleteness, which is manifested as missing sensor readings. Many factors, including sensor failures and/or network disruption, can cause data incompleteness. Furthermore, most IoT systems are severely power-constrained. It is important that we build IoT-based ML systems that are robust against data incompleteness while simultaneously being energy efficient. This paper presents an empirical study of SECOE – a recent technique for alleviating data incompleteness in IoT – with respect to its energy bottlenecks. Towards addressing the energy bottlenecks of SECOE, we propose ENAMLE – a proactive, energy-aware technique for mitigating the impact of concurrent missing data. ENAMLE is unique in the sense that it builds an energy-aware ensemble of sub-models, each trained with a subset of sensors chosen carefully based on their correlations. Furthermore, at inference time, ENAMLE adaptively alters the number of the ensemble of models based on the amount of missing data rate and the energy-accuracy trade-off. ENAMLE's design includes several novel mechanisms for minimizing energy consumption while maintaining accuracy. We present extensive experimental studies on two distinct datasets that demonstrate the energy efficiency of ENAMLE and its ability to alleviate sensor failures.

I. INTRODUCTION

Many areas of human activity have been revolutionized by the Internet of Things (IoT) paradigm, including but not limited to agriculture, healthcare, physical infrastructure management, and manufacturing [1]. Considering that many IoT applications demand real-time response, ML is being increasingly employed for decision-making in IoT ecosystems. Nest Learning Thermostat is a notable example of harnessing the power of ML in IoT. Nest uses ML to learn user preferences and behavior, then adjusts the temperature automatically to optimize energy efficiency, providing users with a comfortable living experience and reducing energy waste and utility bills [2].

There are many challenges in designing robust and effective ML applications for IoT environments. Two major challenges are (a) the possibility of missing data streams during inference time and (b) severe power constraints of IoT devices due to their limited hardware capabilities. Data incompleteness commonly arises due to sensor hardware failure or network connection issues. Such failures during inference time disrupt the operation of the ML-based application and reduce its effectiveness. In particular, in the case of concurrent failures of multiple sensors, the ML application would have to perform predictions/classifications using incomplete data, and this severely degrades its accuracy [3]. In addition, time-critical applications that require a real-time or semi-real-time response are typically deployed on edge devices, such as Raspberry PI or NVIDIA Jetson Nano, rather than the cloud. These devices are severely constrained. This limitation not only affects the throughput of complex ML algorithms but also makes it challenging to run ML algorithms continuously for a long duration.

This paper tackles the problem of data incompleteness due to sensor failures at inference time while simultaneously optimizing the device's energy consumption. Several previous works have attempted to overcome data incompleteness [3,9-12]. The majority of existing related works are focused on providing efficient methods for handling data incompleteness. However, to our knowledge, the existing methods are designed while overlooking the critical aspect of energy consumption, making them unsuitable for IoT resource-constrained devices.

Our research objective is to design an energy-aware approach to make IoT ML applications resilient against data incompleteness during inference. With this aim in mind, we first carry out an in-depth experimental study on SECOE, a recent promising method designed to overcome the limitation of imputation-based techniques against continually missing data from multiple sensors. This study focuses on identifying SECOE's energy consumption hotspots.

In this paper, we propose ENAMLE—an energy-aware, proactive approach that mitigates the impact of missing data through an ensemble [4] of ML models. ENAMLE is specifically built for IoT devices. Our approach includes a unique inference architecture that enables it to mitigate concurrent missing data from multiple sensors while effectively managing energy consumption. ENAMLE's novel design illustrates how alternative ensemble techniques can be harnessed for alleviating sensor failures in resource-constrained devices in IoT systems.

In designing ENAMLE, this paper makes the following technical contributions:

1) We present an extensive experimental study of SECOE operating on an IoT device. We evaluate its performance while focusing mainly on analyzing its energy usage and bottlenecks, demonstrating precisely how SECOE performs on an IoT device and what its trade-offs are with respect to the number of sub-models.

2) ENAMLE enhances the robustness of ML applications on energy-constrained IoT devices through a combination of two unique techniques. First, ENAMLE carefully builds ensembles based on correlations of sensor data streams, with each model in the ensemble trained with a data set that omits a distinct set of sensor values. Second, at the inference phase, ENAMLE adaptively adjusts the number of models in the ensemble in a way that optimizes the trade-off between energy and accuracy based on the missing data rate.

3) We evaluate ENAMLE on an IoT device, using three well-known supervised ML methods, namely, Multi-Layer Perceptron (MLP), Random Forest (RF), and Support-Vector Machine (SVM) [5], on two distinct datasets and compared with existing approaches. Our research findings show that ENAMLE substantially reduces energy consumption and optimizes throughput while maintaining accuracy compared with SECOE, thereby efficiently overcoming sensor failure-induced data incompleteness in IoT.

The remaining sections of the paper are organized as follows. The motivation and background are described in Section II. In Section III, SECOE's examination is conducted in which we analyze its energy profiling. Then, Section IV details the design of ENAMLE, followed by Section V, in which we demonstrate the conducted experiments and results. Then, Section VI describes the related work. Finally, we conclude in Section VII.

## II. MOTIVATION AND BACKGROUND

When building ML models for IoT devices, it is critical to consider their energy constraints. Taking this approach ensures that the operation is sustainable and prolonged while optimizing resource utilization to maximize the efficiency and effectiveness of IoT applications. An exemplary approach is to develop lightweight models. This not only extends the battery life of IoT devices but also enables longer deployments and reduces the necessity of maintenance. Quantization and pruning are two of the most popular techniques for developing light Artificial intelligence (AI) applications to suit resource-constrained devices [15]. Consequently, approaches for IoT ML-based to address data incompleteness should also be implemented to suit resource-constrained devices.

In the same context, missing data at inference time could drastically impact the accuracy of the ML application. Hence, many approaches have been proposed to overcome data incompleteness. Most of these techniques are imputation-based techniques [9-10], which often suffer from the Out-of-Distribution (OOD) problem during the simultaneous failure of multiple sensors. In other words, they generate fillings for the missing sensor data, resulting in a complete sample, which often deviates from the training data distribution, making the ML model misclassify such a sample. In fact, it has been shown in [3] that missing data from multiple sensors has caused a significant drop by 40%-54%, even using an imputation-based approach. Thus, in [3], the authors designed a Sensor Correlation-based Ensemble (SECOE) to overcome these imputation-based approaches. The fundamental purpose of SECOE is to ameliorate the influence of the missing data streams due to sensor failure on the performance of IoT ML-based applications at inference time by avoiding imputation. SECOE is an ensemble of sub-models, each built on a different subset of sensors from the IoT system. SECOE uses sensors' correlation to reduce the number of models in the ensemble and constructs sub-models that mimic the base model's performance – a model trained using all of the system's IoT sensor data. In [3], the paper introduced Sensor Selection for Sub-models (SSMs), a method that selects distinct 50% of sensors from each correlated group of sensors for each sub-model, forming the sub-model features. Before proceeding with the sensor selection process, SSM finds the minimum number of sub-models (*MinM*) needed to ensure each sensor $x$ is included in at least one sub-model and omitted from another sub-model. This ensures that at inference time, if a sensor $x$ fails, then there would always be at least one suitable sub-model that is ready to predict in a real-time way. This suitable sub-model is trained using a subset of sensors, including some correlated sensors to sensor $x$, while excluding sensor $x$. This has made SECOE attain high classification accuracy even though up to 50% of sensors fail, with no need to impute the missing values of these sensors.

Unfortunately, SECOE is not energy-aware. In other words, it has been implemented without taking into consideration that such an approach would run on an IoT resource-constrained device. In contrast to past research, our goal is to design an energy-aware approach to make IoT ML applications resilient against concurrent missing data caused by sensor failures during inference. Our research objective is to proactively minimize the effect of missing data via an ensemble of ML models, each built using a distinct subset of sensors selected carefully from the entire IoT sensors based on their correlation. To the best of our knowledge, none of the existing work offers an efficient and energy-aware approach for handling data incompleteness. In this paper, we present ENAMLE, where, at the inference phase, it adaptively mitigates data incompleteness by adjusting the number of models in the ensemble in a trade-off manner between energy and accuracy based on the missing data rate. ENAMLE's adaptability allows efficient energy management of IoT ML-based applications running on resource-constrained devices while preserving accuracy.

## III. ENERGY PROFILING AND ANALYSIS

This section details the technical components and procedural methodology used to study SECOE's energy consumption on an IoT device, followed by an extensive analysis of the SECOE's performance, focusing on its energy consumption.

## A. Devices

For our purposes, we wanted to use a device to mimic an IoT device that is low-cost and equipped with reasonable hardware. Thus, we have chosen Raspberry Pi 3 Model B+ [16]. To calculate the energy consumption precisely, we have utilized UM25C, a hardware USB multimeter, to monitor and record the real-time consumption of the Current in Amperes (A) and Power in Watts (W) [6].

## B. SECOE's System Component

The internal implementation of SECOE at Inference-time consists of five operations. First, Finding sub-models. This operation searches over the features of sub-models and returns the sub-model(s) with the least matched failed sensors (suitable sub-model(s)). Second, Detecting sub-models. This operation involves determining and preparing which model would make the prediction. For instance, in some cases where only two suitable models exist, the Inference would be handled by the model that achieved the highest training accuracy. The third operation imputes missing data streams when required. Inference is the fourth operation, which includes only the ML models' predictions. Finally, the fifth operation performs majority voting of the models' predictions (ensemble). In this work, we computed the energy consumption for each operation individually and then summed them together, resulting in an accurate computation of the total energy usage of SECOE.

## C. Profiling Experimental Setup

Our experiments were conducted on three datasets from the UCI repository [7]: Dry-Beans (DB), Steel Plates Fault (SPF), and Wall-Following Robot Navigation (WRN). Table 1 shows a summary of the used datasets in this work. As minor preprocessing, the values of the first two datasets were normalized. In this paper, models are all constructed using the default settings and architecture defined by the "Sciket-Learn" Python library [8]. We examine SECOE's on Raspberry Pi using three supervised ML methods, one from each category: neural networks (MLP), ensemble methods (RF), and classical ML (SVM) [5]. All models were trained on 85% of the dataset size and evaluated on the remaining 15%. In this work, we used three metrics for evaluation: prediction accuracy (%) (1), inference throughput (inferences/ms) (2), and energy consumption in millijoules (mJ) (3). Calculating joules in Eq. (3) needs ampere per second (Amps/s) measurement to find W. Note that in our experiment, the Voltage V = 5.039 is the electric potential charging difference.

$$Throughput = \frac{The\ \#\ of\ inferences}{The\ total\ execuation\ time} \quad (1)$$

$$Accuracy = \frac{The\ \#\ of\ correctly\ classified\ lables}{The\ total\ \#\ of\ predictions} \quad (2)$$

$$Joules\ (J) = W \times s, \quad where\ W = V\ x\ A \quad (3)$$

Initially, for each dataset, we compute the above metrics of the base model and SECOE using different ensembles of models during the failure of sensors and analyze their performance. Specifically, considering other metrics than energy consumption, to further study and determine the patterns that would determine a way to find an ensemble of models that could produce good inference accuracy and throughput while consuming less energy, we conduct experiments on the ensemble of size (*MinM*) that SECOE produced for each dataset, as well as ensembles using arbitrary numbers of sub-models greater than the *MinM*. These numbers correspond to the *MinM+2, MinM+4, MinM+8, MinM+12,* and *MinM+16* sub-models. Note that the *MinM* that SECOE found is six models for the SPF dataset and four for the other two datasets. In all experiments, for the sake of simplicity, we use the Mean Imputation approach to replace the values of missing sensors with the mean of their records from the training set.

TABLE 1. SUMMARY OF THE DATASETS.

| Dataset | Features | Number of Samples | Number of Classes |
|---|---|---|---|
| Dry-Beans (DB) | 16 dry beans' geometric features. | ~14K | 7 Bean species. |
| Steel Plates Fault (SPF) | 27 features describe plates' geometric shapes. | ~ 2K | 7 Plates' faults. |
| Wall-Following Robot Navigation (WRN) | 24 features correspond to numerical ultrasound sensors' readings gathered from embedded sensors on a robot. | ~ 5.5K | 4 Robots' movements. |

## D. Profiling Results

To mimic real-world scenarios, we simulated several test cases where sensors failed simultaneously at random. We demonstrate the performance of SECOE using the different number of sub-models mentioned earlier during the random failure of sensors from 5% to 60%. For each percentage of failure, we conducted ten random runs and reported the averaged results. In Fig. 1, we are demonstrating the energy consumption for the SECOE's operations individually to determine the gaps at which SECOE consumes high energy. Fig. 1, as an example, shows the results of one of the examined ensembles, namely *MinM+4*. We can see that the average contribution of the Voting operation toward the total energy usage is 56% and 42% larger than the rest of the operations in the test cases, 5% and 10% of sensor failures, respectively. This is due to the availability of many suitable sub-models, which drove SECOE to consume a lot of energy during the cases of fewer sensor failures (5% and 10%). And, as the sensor failure increases above 10%, the Imputation contributes the most to the total energy, while both the contribution of the Inference and Voting decreases gradually. This is because the more failures, the less suitable sub-models are found. Moreover, Voting becomes the least, along with the other two operations (Finding sub-models and Detecting sub-models). Due to their simple job, the operations Finding and Detecting sub-models consume less energy than the other operations. Furthermore, it is essential to mention that the energy consumption of both Inference and Imputation depends on the type of ML and Imputation methods used.

On the other hand, results in (a) and (b) in Fig. 2 demonstrate the averaged energy and throughput results of 1000 runs using the MLP on the WRN dataset. In general, the results indicate that the more the number of sub-models is, the higher the energy consumption and lower throughput. Overall, with the number of sub-models larger than the *MinM* on all datasets, during the low sensor's failure (e.g., 5%), SECOE required high energy compared to other percentages of failures, causing low inference throughput. This is because the failure of sensors is not high, resulting in the availability of several suitable sub-models (free-of-failure or with least matched faulty sensors), which SECOE utilizes to perform an ensemble of predictions (voting). The *MinM*, on the other hand, has had higher throughput at low sensor failures (5% and 10%) due to the availability of few suitable sub-models. In addition, compared to the *MinM*, both *MinM+2* and *MinM+4* have had almost similar or comparable energy usage and throughput when the failure of sensors exceeds 10%, while *MinM*+8 has gained higher and somewhat close results to *MinM*+2 and *MinM*+4 during 30% or higher failure rates. Note that as the number of failed sensors increases, the *MinM* consumes more energy since it has fewer models and requires more imputation to fill in the missing data from the sensors. Similarly, the base model, even though it consists of one model, consumes significantly more energy when the missing sensor data exceeds 20%. At a failure rate of 60%, the base model consumes about 41% more energy than SECOE when the number of models is *MinM*+4, resulting in a 30% reduction in throughput compared to *MinM*+4. These findings confirm that SECOE is not only more accurate but also better in terms of energy utilization and throughput than the base model during the simultaneous failure of multiple sensors. This is because SECOE reduces the imputation by 50% compared to the base model.

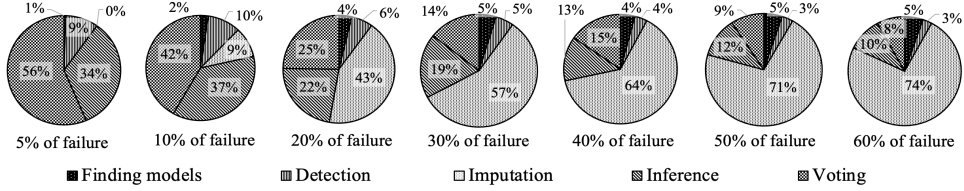

Fig. 1 Energy consumption per operation on the WRN dataset using MLP during random failure of sensors when the number of sub-models is *MinM+4* (8 models).

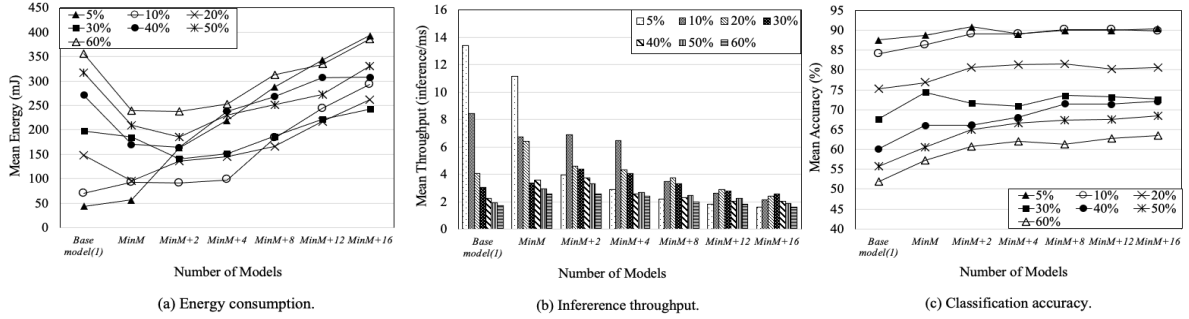

Fig. 2 Mean Energy usage, throughput, and test accuracy of SECOE's different ensembles and Base model during random failure of sensors on the WRN dataset using MLP.

Moreover, according to Fig. 2(c), it has been observed that with a larger ensemble of models than the *MinM*, the accuracy is improved. The larger ensemble of models over *MinM*+4, at failure cases 20%-40%, show better accuracy than other smaller ensembles of models; however, at the failure rate of 50%-60%, their accuracy improvement over the *MinM*+4 is not significant enough to justify the increase in their energy consumption. Thus, based on facts drawn from these results, choosing a specific number of models that achieves high accuracy with reduced energy utilization and maximized throughput should involve a trade-off between energy consumption and accuracy. Unfortunately, SECOE is not an energy-aware approach, and the selection of the number of models is only based on accuracy. Hence, we present in this paper an energy-aware adaptive approach that effectively handles data incompleteness and addresses the energy bottlenecks of SECOE. Note that we do not show figures of the empirical results using the different ML methods for all datasets since they are consistent with the results shown in Fig. 2.

## IV. ENERGY-AWARE ML ENSEMBLE (ENAMLE)

### A. Overview

To make ML-based IoT applications more resilient against data incompleteness due to the failure of sensors, we introduce ENAMLE, a proactive energy-aware ensemble of models with enhanced energy management capabilities to fit resource-constrained IoT devices. ENAMLE demonstrates adaptability in its energy management strategy by dynamically adjusting the number of models used for predictions based on the missing data rate from sensors. This adaptability allows ENAMLE to optimize energy consumption while maintaining acceptable accuracy levels and maximizing the inference throughput. Fig. 3 represents a high overview of our approach. The following subsections describe the architecture of our approach in more detail.

### B. Sub-Models Construction

The training architecture of ENAMLE is inspired by [3]. In particular, in ENAMLE, constructing the sub-models is based on two main components: 1) Forming a correlation between sensors "features" and 2) Forming sensors "features" for each sub-model. The former involves grouping correlated sensors into groups based on their positive Pearson correlation coefficient such that each sensor is only grouped with its highly correlated sensor. After forming the correlated groups, the selection of sensors for each model is accomplished by an iterative sensors selection method, which iteratively selects the leftmost sensors from each correlated group for each sub-model. The amount of selected sensors from each correlated group is equal to 50% of the correlated group. Upon the completion of picking sensors for sub-mode-1, before selecting sensors for the next sub-model, for example, sub-model-2, the correlated groups are shifted left by 1. This step is to ensure that each sensor from each correlated group is included in one model and excluded from another sub-model for more reliability. After iterating over the number of models, our sensor selection method returns sub-models' features. This selection of sensors method enables the sub-models to be trained with different subsets of sensors from each correlated group corresponding to 50% of the entire IoT sensors in the system.

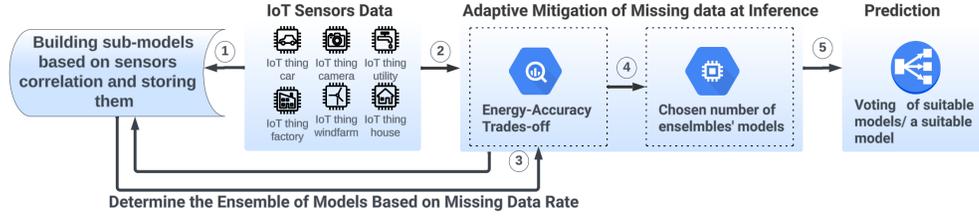

Fig. 3 High overview of the proposed approach.

As a result, during a failure of up to 50% of sensors from each correlated group, our approach could offer a suitable sub-model (free-of-failure) trained on a subset of sensors that are correlated to the faulty ones. This model can make the prediction without imputation and optimizes the accuracy at high missing data rates during inference time. In addition, in cases where no suitable model is free of faulty sensors, the missing data from these sensors will be imputed; however, the imputation is minimized to up to 50% compared to the base model. An example of how our approach builds features for the sub-models is illustrated in Fig. 4.

### C. ENAMLE at Inference

At inference, ENAMLE offers energy-aware resiliency against data incompleteness through Threshold-based model selection (TBMS) and Missing data rates model selection (MDRMS) mechanisms. The TBMS enables restricting the prediction to be done by some of the suitable models, prioritizing energy while considering accuracy. At the same time, the MDRMS allows ENAMLE to utilize different configurations of ensembles, specifically, two configurations, to enable trade-offs between accuracy and energy by prioritizing energy over accuracy at some missing data rates for optimizing the overall system performance. In addition, when data is missing from sensors, in order to optimize accuracy, our method's final prediction is based on the majority of sub-models' votes. However, suppose there are fewer than three suitable sub-models. In that case, the prediction is handled by the most suitable sub-model, which is the model that has the highest accuracy on the training set among the suitable sub-models. The following explains in detail the inference architecture of our approach.

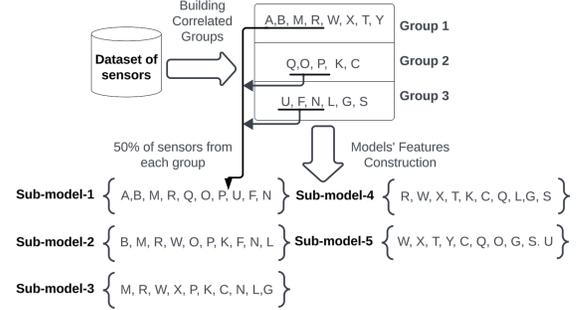

Fig. 4 Example of sub-models construction in ENAMLE.

**TBMS:** ENAMLE introduces a threshold denoted by $t$ that controls the number of available suitable models at inference during sensor failures. This threshold is used to address the energy consumption bottlenecks of SECOE at inference when the failure of sensors is not high (e.g., 5% and 10%). The threshold can be specified as either a percentage or an integer number. By adjusting $t$, ENAMLE can tailor the model selection process, striking a balance between accuracy and energy consumption. For example, suppose we encounter missing data from a sensor $x$, and the found suitable models built without sensor $x$ out of an ensemble of 8 models is six sub-models. In that case, with $t=50\%$, the prediction will be taken care of by the first three models out of the six suitable models. We assigned $t=50\%$ in this work to maximize accuracy as much as possible through voting of the suitable models. A lower value of $t$ might lead to prediction via one sub-model, thereby impacting accuracy.

**MDRMS**: As observed earlier in subsection III.D, selecting one ensemble over another is not always a good choice for all missing data cases if we intend to minimize energy consumption and maintain good accuracy. Thus, to tackle this issue, ENAMLE adapts its model selection strategy based on the missing data rate and utilizes two ensemble configurations ( e.g., *MinM+4 and MinM+8*). In this work, we categorize missing data rates into three ranges: low (5%-10%), moderate (20%-40%), and high (50% and above). At each range, a different ensemble of models is chosen dynamically for prediction by ENAMLE, considering energy consumption, inference accuracy, and throughput. Fig. 5 depicts the workflow of ENAMLE during inference. First, as shown in Fig. 5, in case of encountering missing data, our approach checks the level of the missing rate. If the missing rate is low, it applies the TBMS, restricting the inference to a few suitable models. This is because, at low missing data rates, several suitable models are often available for inference, and the impact of missing data on prediction accuracy is not significant. Thus, by restricting the inference to be performed by some of these models instead of all of them, we enforce the minimization of energy consumption, increasing throughput while maintaining the prediction accuracy. For moderate missing rates, ENAMLE, aiming to improve the robustness of predictions, adapts by utilizing a large ensemble of models. As the missing rate increases, the probability of having suitable models with a small number of matched faulty sensors is higher for a large ensemble configuration than for smaller ones. For example, an ensemble of 12 models might contain models that are free of failure or have a smaller number of matched faulty sensors when 30% of data is missing at inference, while an 8-model ensemble might not. By employing a large ensemble configuration in this missing range, ENAMLE maximizes accuracy while still considering overall energy consumption. With high missing data rates, ENAMLE adjusts its model selection strategy to optimize energy utilization, prioritizing energy consumption. It has been found that selecting a large ensemble of models in this range does not illustrate a noticeable improvement in the prediction accuracy (see subsection III.D). In addition, due to high missing rates, suitable model(s) within the ensemble often would include several matched faulty sensors, raising the necessity to perform more imputations to fill in the missing data before performing inference, leading to high energy consumption. Therefore, at this range, ENAMLE uses an ensemble of models lower than the ones for the moderate missing rates, which balance accuracy and energy consumption and enhance the overall throughput. In Fig. 5, the base model will predict if no missing data exists.

## V. Evaluation

### A. Experimental Setup

Following the experimental setup described in subsection III.C, we evaluate the performance of ENAMLE using the datasets depicted in Table 1. Initially, aiming to optimize the energy and throughput without affecting prediction accuracy, we perform experiments with ENAMLE using two configurations of ensembles that were found to show good results in previous results of profiling SECOE, specifically, *MinM+4* and *MinM+8*. Next, to show how applicable ENAMLE is across different ensembles, we further examine its efficiency with varying combinations of two ensembles of models and provide suggestions on choosing the combination ensemble configuration. To demonstrate the significance of our approach, we compare its performance with SECOE and the base model. Throughout all experiments, we evaluate all approaches based on three metrics: classification accuracy, energy consumption, and inference throughput.

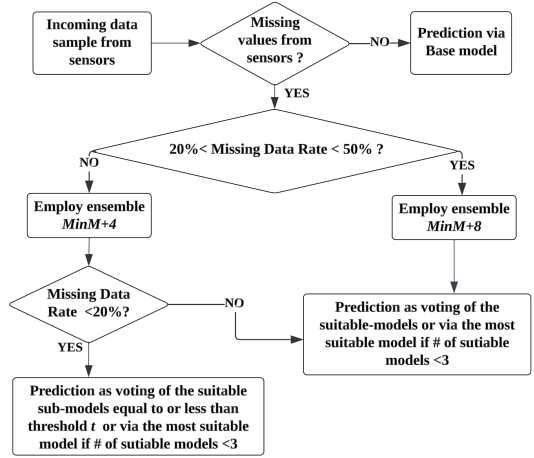

Fig. 5 Example of ENAMLE inference workflow.

### B. Optimization to Higher Accuracy

In this experiment, we examine ENAMLE with the two ensembles *MinM+4* and *MinM+8*. Fig. 6 and Fig. 7 demonstrate the average results of ten random runs per missing data rate from 5%-60%, along with the averaged results across all missing data rates for the WRN and SPF datasets, respectively, using MLP and SVM. The results in Fig. 6(c) and 7(c) indicate that ENAMLE efficiently mitigates the impact of missing data on the accuracy as the rate of the missing data of sensors increases, where at 60%, its accuracy reaches 62.0% and 50.6%, which is better than the base model by 19.8% and 14.0% for both datasets, respectively. In contrast, SECOE shows great performance in accuracy optimization; however, according to Fig. 6(a), ENAMLE significantly reduced the averaged energy consumption across the different missing data rates by 7.7%, 25.6%, and 12.3% compared to SECOE-*MinM+4,* SECOE-*MinM+8*, and Base model, respectively. It also optimizes the averaged inference throughput by 35.7% and 77.1% compared to SECOE-*MinM+4* and SECOE-*MinM+8*, respectively. See Fig. 6(b). Similarly, in Fig. 7(a), ENAMLE cuts the energy by 19.6% and 38.5% less than SECOE-*MinM+4* and SECOE-*MinM+8*, respectively. According to Fig. 7(b), this reduction in energy leads to a 26.8% and 61.4% rise in throughput, more than what has been achieved by SECOE with *MinM+4* and *MinM+8*, respectively. In this dataset, the base model still consumes less energy; however, the energy consumption of ENAMLE becomes very close to that of the base model, resulting in only 6.6% less throughput than the base model. These results confirm the intuition behind ENAMLE: By employing two ensemble configurations along with a threshold mechanism, we can achieve high accuracy with optimized energy consumption and throughput during concurrent missing data due to sensor failures, making it suitable for resource-constrained devices, thereby enhancing the resiliency of ML-based IoT applications against data incompleteness in resource-constrained devices as well as reducing its operation cost.

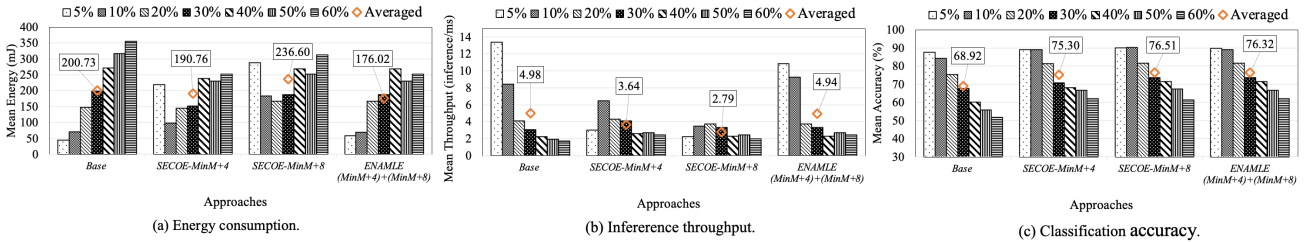

Fig. 6 Performance comparison on the WRN dataset using MLP during different missing data rates.

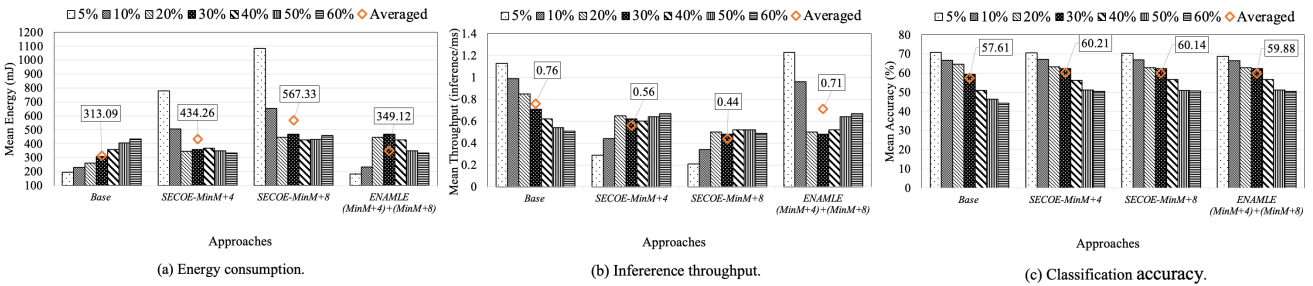

Fig. 7 Performance comparison on the SPF dataset using SVM during different missing data rates.

### C. Effect of Combination of Lower Ensembles

To further validate ENAMLE, we ran several experiments in which we used different combinations of ensemble configurations lower than *MinM+8* on the WRN dataset using the MLP method. Here, we are trying to test the generality of our approach across different ensembles with the goal of reducing energy consumption substantially. Overall, the results from Fig. 8 align with the results from Fig. 6, proving that ENAMLE could offer more energy-efficient models that are robust toward missing data at inference. As shown in (a) and (c) in Fig. 8, using a combination of *MinM+2* for low and high missing data rates and *MinM+4* for moderate missing data rates provides a significant decrease in averaged energy compared with the individual ensembles by SECOE. This energy-saving is achieved with an unnoticeable minor trade-off of less than 0.3% in the averaged accuracy when compared to SECOE-*MinM+4*. The overall average energy usage reached 159.98mJ, which is almost similar to what has been consumed by SECOE-*MinM+2*, and it is 16.1% less in comparison to SECOE-*MinM+4*.

On the other hand, adaptively using both (*MinM* and *MinM+2*) ensembles during inference enabled ENAMLE to consume an average energy of 142.93mJ. This is less than what has been gained by SECOE-*MinM+2* and SECOE-*MinM*. ENAMLE, according to Fig. 8(b), boosted the throughput by enforcing energy minimization. Compared to SECOE-*MinM+8* from Fig. 6(c), an extensive reduction in energy by ENAMLE coupling (*MinM* and *MinM+2*) causes a loss in the averaged accuracy of about 4.5%. However, its energy consumption, based on Fig. 6(a) and Fig.8(a), reached approximately 40.0% less than SECOE-*MinM+8*. In addition, it also has less average energy usage than the base model of nearly 29.0% less, with much better accuracy when missing data is 30% or above. From analyzing these results, we can conclude that the adaptive switching between two different ensembles, combined with the TMBS technique, alleviates the impact of concurrent sensor failures while efficiently optimizing energy. Furthermore, generally, for better results, we suggest using the two ensembles (*MinM+4* and *MinM+8*) in ENAMLE; however, if the primary objective is to significantly lower the overall energy of the application to make it more suitable for very resource-constrained IoT devices, we recommend using (*MinM* and *MinM+2*).

### VI. RELATED WORK

Over recent years, research has aimed to address data incompleteness. Many of the recent promising approaches are AI-based. Some of which are cluster-based. As a recent example, Karmitsa *et al.* [9] proposed a clusterwise linear regression method for processing data incompleteness. Their approach integrates linear regression and clustering to approximate missing data using the available data points that are somewhat similar to the missing ones. Compared to existing work, using linear regression within each cluster precisely predicts the missing data when missing data is less than 25%. Agbo *et al.* [10] presented a method based on data clustering and a robust selection of suitable imputations for each missing variable within each sample. After being tested when 3%-15% of the data were missing, they found that their method performs similarly to the Local Similarity Imputation (LSI) method, but its complexity is less than LSI. On the other hand, some work employed Deep Neural Networks (DNNs) for constructing efficient imputation for missing data. Che et al. [11] offered a modified Recurrent NN (RNN) model using, namely, a Gated Recurrent Unit (GRU) that predicts the future missing values of multivariate time-series data type using historical data and learning future information of the time-series. Zha et al. [12] proposed a self-supervised generative model. It generates time series data to fill the missing instances by relying on temporal correlation. During training, segments of the input time series are randomly masked, and the entire series is extrapolated based on the information learned from unmasked segments. The model outperforms popular imputation methods like Mean. Unlike the above imputation-based approaches, in [3], the authors presented SECOE, a proactive ensemble-based approach to handle data incompleteness at inference time through sub-models, each built with a distinct subset of sensors based on the correlation of sensors. SECOE enables inference without the need to impute the missing data, even at up to 50% of missing data from sensors, exhibiting excellent performance against missing data from sensors.

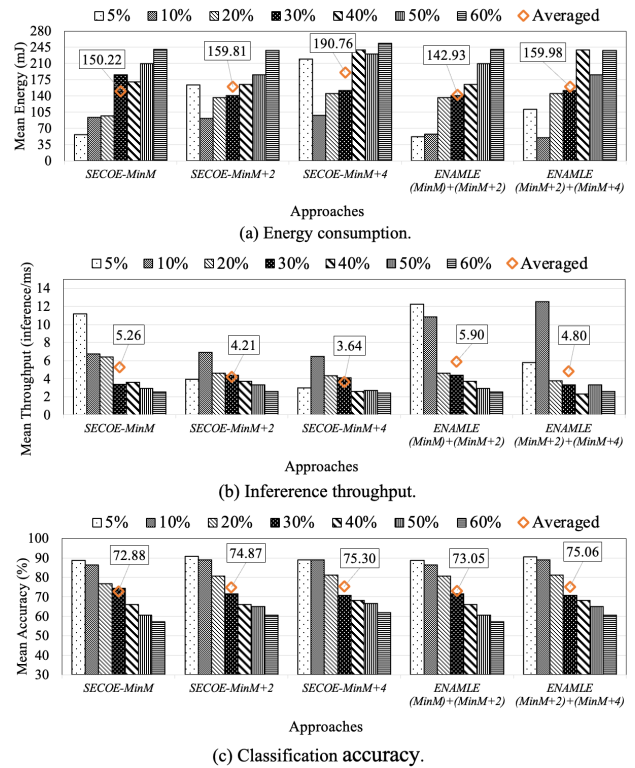

Fig. 8 ENAMLE vs. SECOE using smaller ensembles on the WRN dataset.

Unfortunately, the above AI-based approaches are not energy-aware, limiting their useability in IoT resource-constrained devices. In the same context, throughout the preceding years, several researchers have attempted to address issues in developing AI applications to suit resource-constrained devices. Some of these works concentrate on making the ML network lighter. As a prominent example, MobileNet uses depthwise and pointwise convolutions, with a scaling factor to control the channels of each layer, making it adaptive to different computational budgets while maintaining a trade-off between accuracy and complexity [13]. EfficientNet balances model size and performance through compound scaling [14]. Quantization and model pruning have also been utilized in the previous work. For example, Prakash et al. [15] reduced the size of the model and communication bandwidth while maintaining accuracy through model pruning and quantization in integration with federated learning (FL).

In this paper, motivated by prior research, we aim to provide an efficient energy-aware strategy for mitigating the influence of data incompleteness at inference time on IoT ML-based systems running in IoT resource-constrained devices.

## VII. CONCLUSION

Existing techniques for handling data incompleteness lack energy awareness, limiting their usability in resource-constrained IoT ecosystems. In this paper, after identifying the energy bottlenecks within SECOE, we introduced ENAMLE, an efficient and energy-aware for mitigating the effect of concurrent missing data consisting of an ensemble of sub-models, each trained using a subset of sensors selected based on sensor correlation. During inference, ENAMLE dynamically adjusts the number of models in the ensemble, taking into account the missing data rate and the trade-off between energy consumption and accuracy. At missing data rates from 5% to 60%, ENAMLE effectively maintains the prediction accuracy and optimizes both the energy consumption and throughput. Our approach substantially outperforms SECOE in terms of optimization in the average throughput and energy consumption on all datasets. It also shows less energy than the base model with significantly higher accuracy during high rates of missing data on all datasets. We also studied the performance of ENAMLE using lower numbers of ensemble models and provided suggestions for choosing both for better performance.

REFERENCES

[1] L. Atzori, A. Iera, and G. Morabito, "The internet of things: a survey," *Computer Networks: The Int. J. of Comput. and Telecommun. Netw.*, vol. 54, no. 15, pp. 2787–2805, Oct. 2010.

[2] Greatest of All TOM, "IOT and me: Nest, machine learning, and the Smart Device Revolution," *Technology and Operations Management*. [Online]. Available: https://d3.harvard.edu/platform-rctom/submission/iot-and-me-nest-machine-learning-and-the-smart-device-revolution/. [Accessed: 14-Jan-2024].

[3] Y. AlShehri and L. Ramaswamy, "SECOE: Alleviating Sensors Failure in Machine Learning-Coupled IoT Systems," 2022 21st IEEE Int. Conf. on Mach. Learn. and Applications (ICMLA), Nassau, Bahamas, 2022, pp. 743-747, doi: 10.1109/ICMLA55696.2022.00124.

[4] O. Sagi and L. Rokach, Ensemble learning: A survey,Data Mining and Knowledge Discovery, vol. 8, no. 4,pp. 1–18, 2018.

[5] A. Singh, N. Thakur and A. Sharma, "A review of supervised machine learning algorithms," 2016 3rd Int. Conf. on Comput. for Sustainable Global Develop. (INDIACom), 2016, pp. 1310-1315.

[6] "Rd Um25c Type-c Usb 2.0 Color Lcd Display Dc Tester Voltmeter Ammeter Current Tester Cable Resistance Voltage Meter," *www.alibaba.com*. Available: https://www.alibaba.com/product-detail/RD-UM25C-Type-C-USB-2_60776977613.html?spm=a2700.shop_plgr.41413.14.5b815e850or84Z . [Accessed: 14-Jan-2024].

[7] D. Due and C. Graff, UCI Machine Learning Repository, Irvine, CA: UC, School of Information and Computer Science, 2017. Accessed on: Oct 15, 2021. [Online]. Available: http://archive.ics.uci.edu/ml. [Accessed: 14-Jan-2024].

[8] F. Pedregosa *et al.*, "Scikit-learn: machine learning in python," *The J. of Mach. Learn. Research*, vol 12, pp. 2825-2830, Jan. 2011.

[9] N. Karmitsa, S. Taheri, A. Bagirov and P. Mäkinen, "Missing Value Imputation via Clusterwise Linear Regression," in IEEE Transactions on Knowledge and Data Engineering, vol. 34, no. 4, pp. 1889-1901, 1 April 2022, doi: 10.1109/TKDE.2020.3001694.

[10] Agbo, B., Qin, Y., & Hill, R, "Best fit missing value imputation (BFMVI) Algorithm for Incomplete Data in the Internet of Things," presented at 5th Inter. Conf. on IoT, Big Data and Security (IoTBDS), Prague, Czech, May. 2020, pp. 130-137.

[11] Z. Che, S. Purushotham, K. Cho, D. Sontag, and Y. Liu, "Recurrent neural networks formultivariate time series with missing values," *Sci. Rep.*, vol. 8, no. 1, pp. 1–12, 2018.

[12] M. Zha, S. Wong, M. Liu, T. Zhang, and K. Chen, "Time Series Generation with Masked Autoencoder," arXiv (Cornell University), Jan. 2022, doi: 10.48550/arxiv.2201.07006.

[13] A. G. Howard, M. Zhu, B. Chen, D. Kalenichenko, W. Wang, T. Weyand, M. Andreetto, and H. Adam, "Mobilenets: Efficient convolutional neural networks for mobile vision applications," arXiv preprint arXiv:1704.04861, 2017.

[14] M. Tan and Q. Le, "EfficientNet: Rethinking model scaling for convolutional neural networks," in Proc. Int. Conf. Mach. Learn., 2019, pp. 6105–6114.

[15] P. Prakash et al., "IoT Device Friendly and Communication-Efficient Federated Learn. via Joint Model Pruning and Quantization," in IEEE IoT J., vol. 9, no. 15, pp. 13638-13650, 1 Aug.1, 2022, doi: 10.1109/JIOT.2022.3145865.

[16] Raspberry Pi, "Buy A raspberry pi 3 model B+," *Raspberry Pi*. [Online]. Available: https://www.raspberrypi.com/products/raspberry-pi-3-model-b-plus/. [Accessed: 14-Jan-2024].